\documentclass[11pt,a4paper]{article}

\usepackage{fontspec}
\usepackage{xeCJK}
\usepackage{unicode-math}

\usepackage{amsmath}
\usepackage{booktabs}
\usepackage{graphicx}
\usepackage{float}
\usepackage{array}
\usepackage{microtype}
\usepackage[preprint]{acl}

\usepackage{arydshln}

\usepackage{colortbl}
\definecolor{initialbg}{HTML}{DBE5F1}  %
\definecolor{finalbg}{HTML}{FFF2CC}    %
\definecolor{tonebg}{HTML}{FBE4E4}     %
\definecolor{initialfg}{HTML}{1F4E79}
\definecolor{finalfg}{HTML}{BF8F00}
\definecolor{tonefg}{HTML}{C00000}

\setCJKsansfont{FandolHei-Regular.otf}[BoldFont=FandolHei-Bold.otf]
\setCJKmonofont{FandolFang-Regular.otf}
\newfontfamily\ipafont{CharisSIL-Regular.ttf}[BoldFont=CharisSIL-Bold.ttf,ItalicFont=CharisSIL-Italic.ttf,BoldItalicFont=CharisSIL-BoldItalic.ttf]
\newcommand{\ipa}[1]{{\ipafont #1}}

\newcommand{\initial}{\textsc{Initial}}
\newcommand{\final}{\textsc{Final}}
\newcommand{\tone}{\textsc{Tone}}
\newcommand{\layer}[1]{L#1}
\newcommand{\orange}[1]{\textcolor{black}{#1}}

\newcommand{\hlinitial}[1]{\textcolor{initialfg}{\textbf{#1}}}  %
\newcommand{\hlfinal}[1]{\textcolor{finalfg}{\textbf{#1}}}      %
\newcommand{\hltone}[1]{\textcolor{tonefg}{\textbf{#1}}}        %

\newcommand{\behavsame}{\(\checkmark\)}
\newcommand{\behavdiff}{\(\times\)}
\newcommand{\behavacc}[1]{#1}
\newcommand{\behavpy}[3]{\textit{\textcolor{initialfg}{#1}\textcolor{finalfg}{#2}\textcolor{tonefg}{#3}}}

\title{Phonological Knowledge of Chinese Characters in Language Models}

\author{
Xiaozhe Ji$^{1}$ \quad Renfen Hu$^{2}$ \quad Hai Hu$^{3}$\thanks{Corresponding author.}\\[3pt]
  $^{1}$\small  School of Foreign Languages and Literature, Beijing Normal University\\
  $^{2}$\small  School of International Chinese Language Education, Beijing Normal University\\
  $^{3}$\small Department of Language Science and Technology, Division of Artificial Intelligence and the Humanities, \\
  \small The Hong Kong Polytechnic University\\[1pt]
  \small \texttt{\{202521100023, irishu\}@mail.bnu.edu.cn} \quad \texttt{hai.hu@polyu.edu.hk}}

\date{}

\hypersetup{
  colorlinks=true,
  linkcolor=blue
}
\usepackage{tikz}

\newcommand{\tonecurve}[1]{%
  \begin{tikzpicture}[x=0.75cm,y=0.35cm,baseline=-0.5ex]
    \path[use as bounding box] (0,0) rectangle (1,1);
    \draw[tonefg,line width=0.8pt,line cap=round]
      plot coordinates {#1};
  \end{tikzpicture}%
}
\begin{document}
\flushbottom
\maketitle

\begin{abstract}
How much phonological knowledge do language models (LMs) possess about Chinese characters, and how is this knowledge represented and used? We introduce \textbf{PinyinBench}, a controlled benchmark covering homophones, phonological features (initials, finals, and tones), and polyphonic characters. We investigate 17 text-only and multimodal LMs through three complementary approaches: behavioral evaluation, layer-wise probing, and activation steering. 
Behaviorally, models demonstrate substantial knowledge of initials and finals, with a frontier model (DS-v4-flash) scoring 90\%+ in homophone judgment, when two characters share no phonological features. However, its accuracy drops sharply to 72\% when two characters differ only in lexical tone. 
Probing reveals that tone is highly decodable from embeddings and improves modestly across layers, whereas segmental information starts less decodable and improves substantially with depth.
Activation steering shows that segmental representations can systematically influence generation, while tone exhibits much weaker effects. Furthermore, audio-bearing training improves phonological decodability, including  tone, without corresponding improvements in steering effectiveness. 
Together, these findings reveal distinct representational and functional profiles for segmental and tonal information, with tonal information much harder to access and manipulate, demonstrating the necessity for combined evaluation, probing, and causal intervention for understanding linguistic knowledge in LMs.

\end{abstract}

\section{Introduction}
\label{sec:intro}

Recent advances in language modeling have led to impressive performance in mathematical problem
solving, reasoning, and code generation \citep{wang-etal-2024-math,
muennighoff-etal-2025-s1,
lei-etal-2025-planning}. However, the phonological knowledge of language models (LMs) remains relatively understudied. While recent work has evaluated phonological abilities \citep{suvarna2024phonologybench,yue2026phunbench} and probed internal phonological representations \citep{liao-shi-2026-tokenization}, it remains unclear how phonological knowledge is internally represented and whether these representations are functionally accessible.
It is also unknown how phonological knowledge differs between text-only LMs and models that have additionally been trained on audio.

Chinese offers an interesting test case because its logographic writing system in general lacks systematic grapheme-to-phoneme correspondences, which are prevalent in alphabetic  languages such as English or German.\footnote{For some characters though, phonetic components provide partial pronunciation cues \citep{shu2003properties}.} 
As illustrated in Table~\ref{tab:syllable-decomposition},
a Mandarin syllable consists of an optional initial,
a final, and a tone.
Initials and finals describe its segmental structure,
whereas tone is a suprasegmental feature specifying
the pitch over the syllable.
Changing the tone can distinguish words with the
same initial and final.

\begin{table}[t]
\centering
\small
\setlength{\tabcolsep}{3pt}
\renewcommand{\arraystretch}{1.15}

\resizebox{\columnwidth}{!}{%
\begin{tabular}{lcccc}
\toprule
& & \multicolumn{2}{c}{\textbf{Segmental}}
  & \textbf{Suprasegmental} \\
\cmidrule(lr){3-4}\cmidrule(lr){5-5}
& \textbf{Full syllable}
& \textcolor{initialfg}{\textbf{Initial}}
& \textcolor{finalfg}{\textbf{Final}}
& \textcolor{tonefg}{\textbf{Tone}} \\
\midrule
\textbf{Pinyin}
& \textit{bai3 / bǎi}
& \textcolor{initialfg}{b}
& \textcolor{finalfg}{ai}
& \textcolor{tonefg}{3 / ˇ} \\
\textbf{IPA}
& {\ipafont [paɪ̯˨˩˦]}
& {\ipafont\textcolor{initialfg}{[p]}}
& {\ipafont\textcolor{finalfg}{[aɪ̯]}}
& {\ipafont\textcolor{tonefg}{[˨˩˦]}} \\
\textbf{Character}
& 摆 & -- & -- & -- \\
\midrule
\textbf{Mandarin inventory}
& -- 
& \textcolor{initialfg}{21}
& \textcolor{finalfg}{39}
& \textcolor{tonefg}{4} \\
\bottomrule
\end{tabular}%
}

\medskip
{\footnotesize
\textbf{Same initial and final, different tones and meanings}
\par}
\smallskip

\begin{tabular*}{\columnwidth}
{@{\extracolsep{\fill}}cccc@{}}
\textcolor{tonefg}{\textbf{bāi}}
& \textcolor{tonefg}{\textbf{bái}}
& \textcolor{tonefg}{\textbf{bǎi}}
& \textcolor{tonefg}{\textbf{bài}} \\
掰 & 白 & 摆 & 拜 \\
{\scriptsize (break apart)}
& {\scriptsize (white)}
& {\scriptsize (put)}
& {\scriptsize (bow)} \\
\tonecurve{(0,0.9) (1,0.9)}
& \tonecurve{(0,0.4) (1,0.9)}
& \tonecurve{(0,0.4) (0.4,0.05) (1,0.75)}
& \tonecurve{(0,0.9) (1,0.05)} \\
{\scriptsize 1: High-level}
& {\scriptsize 2: Rising}
& {\scriptsize 3: Falling--rising}
& {\scriptsize 4: Falling} \\
\bottomrule
\end{tabular*}

\caption{
Mandarin syllable structure.
Curves show schematic pitch contours in isolation;
neutral tone is not illustrated.
}
\label{tab:syllable-decomposition}
  \vspace{-2mm}
\end{table}

In this paper, we introduce \textbf{PinyinBench},
a controlled benchmark comprising 38,183 minimal
pairs for evaluating LMs' phonological knowledge
of Chinese characters through three tasks:
homophone judgment, feature judgment, and
polyphone disambiguation. Beyond behavioral evaluation with PinyinBench, we investigate the internal representation and functional accessibility of phonological knowledge. Specifically, we ask the following questions for Chinese:
\textbf{(RQ1)} To what extent do text-only and multimodal LMs possess phonological knowledge?
\textbf{(RQ2)} Where are different phonological dimensions linearly decodable?
\textbf{(RQ3)} Can these representations be manipulated to influence model outputs?

We conduct three linked experiments:
\emph{(i) Behavioral evaluation} using controlled phonological judgments from PinyinBench;
\emph{(ii) Layer-wise probing} of initials, finals, and tones;
and \emph{(iii) Activation steering} to test whether phonological features can be manipulated to influence generation.

Across three experiments, we find that LMs possess substantial phonological knowledge of Chinese characters, but this knowledge is uneven.
\emph{Behaviorally}, models reliably distinguish initials and finals but struggle with tones; text-only base models perform near chance ($51$--$57\%$) on tone-only contrasts.
\emph{Probing} shows that tone is highly decodable from character embeddings, whereas segmental information becomes increasingly decodable in deeper layers.
\emph{Steering} reveals that segmental representations can systematically influence generation, while tone is much less responsive. Audio-bearing training improves decodability but not steering effectiveness.

These findings reveal a fundamental dissociation between phonological knowledge being encoded and being functionally accessible. They also suggest distinct representational and functional profiles for segmental and tonal information, demonstrating why behavioral evaluation, probing, and causal intervention are all necessary for understanding linguistic knowledge in LMs.

\begin{table*}[t]
  \centering
  \small
  \renewcommand{\arraystretch}{1.05}
  \setlength{\tabcolsep}{5pt}
  \begin{tabular}{@{}>{\raggedright\arraybackslash}l l l l l@{}}
  \toprule
  \textbf{Task \& Prompt} & \textbf{Target} & \textbf{Positive} & \textbf{Distractor} & \textbf{Condition} \\
  \midrule
  \textbf{Task 1: Homophone Judgment} & & & 歹 (\hlinitial{d}ǎi) & \hlinitial{Initial-mismatch} \\
  “A”和“B”的声母、韵母和声调完全相同 & 摆 (bǎi) & 百 (bǎi) & 比 (b\hlfinal{ǐ}) & \hlfinal{Final-mismatch} \\
  (A and B have identical initial, final \& tone) & & & 白 (bái-\hltone{tone2}) & \hltone{Tone-mismatch} \\
  \cmidrule{1-5}
  \textbf{Task 2: Feature Judgment} & & & 歹 (\hlinitial{d}ǎi) & \hlinitial{initial} \\
  “A”和“B”声母/韵母/声调相同 & 摆 (bǎi) & 百 (bǎi) & 比 (b\hlfinal{ǐ}) & \hlfinal{final} \\
  (A and B share the same initial / final / tone) & & & 白 (bái-\hltone{tone2}) & \hltone{tone} \\
  \bottomrule
  \end{tabular}
  \caption{
  \orange{Example sentence templates for homophone judgment and feature judgment tasks in PinyinBench. Since PinyinBench is a minimal pair benchmark, we have Target and Positive characters in the good sentence, and Target and Distractor in the bad sentence. }
  The target is 摆 (bǎi); the positive candidate is 百 (bǎi).
  Each illustrated distractor differs from the target in one dimension,
  highlighted in blue (initial), yellow (final), or red (tone).
  Homophone judgment requires a match on all three dimensions;
  feature judgment specifies one dimension in the prompt.
  The remaining homophone judgment conditions are defined in
  \S\ref{sec:setup-bench}.
  Task 3 is represented in Table~\ref{tab:pinyinbench-polyphone} in Appendix~\ref{app:class3}.
  }
  \label{tab:pinyinbench-schematic}
   \vspace{-2mm}
\end{table*}

\section{Related Work}
\label{sec:related}

\paragraph{Phonological evaluation of language models.}
\orange{A number of benchmarks have been constructed recently to evaluate knowledge in phonology in English and Chinese. }
PhonologyBench established phonological competence as a benchmark-worthy capability,
showing that English LLMs handle grapheme-to-phoneme conversion, syllable counting, and
rhyme generation well above chance but $17$--$45\%$ below humans
\citep{suvarna2024phonologybench}; P-CoT later showed much of this latent ability can be
elicited with structured prompting \citep{jang2025pcot}. 
The closest prior efforts in Chinese probe homophones, and they find that models lean on memorized internet substitution
patterns rather than productive phonological reasoning, collapsing on controlled or
low-frequency items \citep{ma2025phonothink,ma2025reasoning}. 
\citet{yue2026phunbench} introduce Phun-Bench, a Chinese benchmark spanning homophony,
rhyme, and phonetic similarity, and reach a closely related behavioral conclusion---models
recall correct pronunciations but fail to wield phonological knowledge as flexibly as human
speakers---which they connect to a hypothesized perception mechanism. 
\orange{Our proposed PinyinBench is the first one that isolates initials, finals, and tones as independent
variables under fixed templates and calibrated frequency.
We target phonological knowledge of commonly used Chinese characters rather than internet homophone memes.}

\paragraph{Probing and representational geometry.}
Probing examines what linguistic information can be recovered
from a model's internal representations by training a classifier
to predict linguistic labels from those representations.
Probe accuracy, however, can reflect both information in the
representations and the probe's capacity to learn the task.
To distinguish these contributions, \citet{hewitt2019control}
introduce control tasks and selectivity, comparing performance
on linguistic labels with performance on randomized labels.
A complementary approach examines representational differences
elicited by linguistic minimal pairs.
\citet{chen2025minimalpairs} use these activation differences
to characterize linguistic similarity across language models.
We adopt linear probing with control tasks to examine how
initial, final, and tone information is represented across
model layers and prompt conditions.

\paragraph{Activation steering and causal intervention.}
Activation steering modifies a model's internal activations
during inference to influence its outputs without updating
its parameters.
Such interventions allow researchers to examine how changes
in internal representations affect model behavior.
\citet{rimsky2024caa} derive steering directions from
contrasting examples of behavior and show that adding these
directions to model activations can guide the responses
of Llama~2.
In the phonological domain, \citet{mclaughlin2025rhyme}
identify phoneme-related directions using linear probes
and intervene on token embeddings to influence English
rhyming behavior in Llama-3.2-1B-Instruct.
We adopt activation steering to examine how interventions
on representations of Mandarin initials, finals, and tones
affect model output probabilities.

\section{The PinyinBench and General Experimental Setup}
\label{sec:benchmark}

PinyinBench \orange{is a minimal-pair benchmark that} evaluates knowledge of Chinese character pronunciation through
three tasks: \emph{homophone judgment}, \emph{feature judgment}, and
\emph{polyphone disambiguation}. Tables~\ref{tab:pinyinbench-schematic}
and~\ref{tab:pinyinbench-polyphone} illustrate the prompts and example
items. Experimental items are constructed from Chinese character
lists and their Pinyin pronunciations.
Using the 3,500 most common characters, we construct an
evaluation set comprising 12,600 minimal pairs for homophone
judgment, 5,400 for feature judgment, and 5,000 for polyphone
disambiguation.
To examine performance on less frequent characters, we
add 11,142 homophone judgment pairs and 4,041 feature
judgment pairs drawn from the 3,500--7,000 character band.
The resulting evaluation set contains 38,183 minimal pairs.

Experiment~1 evaluates performance on these tasks; Experiments~2
and~3 draw on the benchmark's character resources and phonological labels
for probing and steering.

Character pronunciations are obtained using
\texttt{pypinyin} and converted to IPA using
\texttt{Dragonmapper}.
We characterize each pronunciation by its initial,
final, and lexical tone (tones 1--4), deriving initial
and final labels from the IPA transcription
(Appendix~\ref{app:ipa-split}).
Homophone judgment and feature judgment use only
monophonic characters to avoid pronunciation ambiguity,
whereas polyphone disambiguation uses characters with
multiple context-dependent readings.

\subsection{Task Construction}
\label{sec:setup-bench}
\orange{As shown in Tables~\ref{tab:pinyinbench-schematic} and \ref{tab:pinyinbench-polyphone}, 
each item contains a correct and an incorrect sentence about a
\emph{target character}.
}
 The correct sentence pairs the target with a
\emph{positive candidate} that satisfies the stated phonological
relation. The incorrect sentence uses a \emph{distractor} that violates
that relation. The three tasks differ in the relation being judged
and in how the distractor is constructed. The first two tasks use monophonic characters, while the last uses polyphonic characters.

\paragraph{Task 1: Homophone judgment.}
The sentence asserts that the target and candidate have identical
initials, finals, and tones. The positive candidate is a true homophone
of the target. Seven conditions specify the distractor's phonological
overlap with the target. In the three \emph{mismatch} conditions,
\emph{initial-mismatch}, \emph{final-mismatch}, and \emph{tone-mismatch},
the named dimension differs while the other two match. In the three
\emph{match} conditions, \emph{initial-match}, \emph{final-match}, and
\emph{tone-match}, the named dimension matches while the other two
differ. In the \emph{disjoint} condition, all three dimensions differ.
Table~\ref{tab:pinyinbench-schematic} illustrates the three
mismatch conditions.

\paragraph{Task 2: Feature judgment.}
The sentence explicitly asks whether the target and candidate share
one specified dimension: initial, final, or tone (see Table~\ref{tab:pinyinbench-schematic}). The positive
candidate shares the named dimension, while the distractor differs
on it. This task tests the same dimensions as homophone judgment
but names the dimension to be judged in the prompt.

\paragraph{Task 3: Polyphone disambiguation.}
This task tests whether models can identify the correct pronunciation
of a polyphonic character in a given word. The model judges
whether the target, as pronounced in that word, has the same
pronunciation as a candidate character
(see Table~\ref{tab:pinyinbench-polyphone} in Appendix~\ref{app:class3}).
For example, 长 in 生长 (Eng: \textit{grow}) is pronounced \textit{zhǎng}, so the
positive candidate 涨 (\textit{zhǎng}) matches this reading; \orange{however the same character 长 is read \textit{cháng} in 长江 (Eng: \textit{Yangtze river}). We thus construct minimal pair items to examine whether the model can select the correct reading based on the context. }
\orange{For space reasons, we present details of this task and its results in Appendix~\ref{app:class3}. }

\subsection{Models}
\label{sec:setup-models}
The behavioral study spans fifteen Qwen base checkpoints across three generations
(Qwen2.5~\citep{qwen2024qwen25}, Qwen3~\citep{qwen2025qwen3},
Qwen3.5\footnote{Qwen3.5 checkpoints are the publicly released weights on Hugging
Face; no technical report was available at the time of writing.}), from
$0.5$B to $32$B, plus Qwen2.5-Omni-7B~\citep{xu2025qwenomni}
and DeepSeek-V4-Flash~\citep{deepseekai2026v4}
(the \texttt{deepseek-chat} endpoint, non-thinking mode) as a frontier reference model accessed via its API. The representational analyses center on the Qwen2.5 base family
($0.5/1.5/3/7$B) for a clean scaling axis and add Qwen2.5-Omni-7B~\citep{xu2025qwenomni}
(multimodal) for the modality comparison. For activation steering, we use Qwen2.5-7B-base and
Qwen2.5-Omni-7B. For each model, we construct steering
directions from its own hidden states and apply them
to that same model, allowing us to compare how the two
models respond to phonological interventions.

\paragraph{Scoring.}
\label{sec:benchmark-scoring}
All models are evaluated zero-shot.
For Qwen base models, an item is correct if the
correct sentence receives a higher total
log-probability than the incorrect sentence.
For Qwen2.5-Omni-7B and DeepSeek-V4-Flash, we use
binary-choice prompts and score the returned
option letter against the correct answer.
Chance accuracy is 50\% in both formats.

\section{Experiment 1: Behavioral Discrimination}
\label{sec:exp1}
This experiment addresses the paper's first research question:
do text-based and multimodal language models have phonological
knowledge of Chinese characters? We examine two sub-questions. For monophonic characters,
\textbf{RQ1.1:} Do models discriminate initials, finals, and tones, and how well on each
dimension?
\textbf{RQ1.2:} How does their performance vary with phonological
distance and lexical frequency?
We first evaluate models on controlled homophone minimal pairs
(\S\ref{sec:behavioral}), then examine each of these factors.
We also evaluate polyphone disambiguation
(Appendix~\ref{app:class3}).

\subsection{Behavioral Results (RQ1.1)}
\label{sec:behavioral}

Table~\ref{tab:behav-results} shows the performance of
Qwen2.5 series and DeepSeek-V4 on our homophone minimal
pair benchmark in different conditions.
Model performance generally improves as more phonological
dimensions (initial, final, and tone) differ between
the target and distractor.
When the target and distractor differ in all three
dimensions (condition: disjoint), Qwen2.5 base models
achieve accuracies of 56.3\%--78.7\%.
Most models achieve their highest accuracy in this condition.
When the target and distractor differ only in tone
(condition: tone-mismatch), accuracy drops to
51.1\%--55.7\% across Qwen2.5 base models,
remaining close to the 50\% chance level even at 32B.
DeepSeek-V4 achieves over 90\% accuracy in every condition
except tone-mismatch, where it scores 72.7\%,
indicating that tonal contrasts remain challenging
even for this stronger model.
Qwen2.5-Omni-7B outperforms Qwen2.5-7B-base across
all conditions but remains below DeepSeek-V4.
A similar pattern is observed in feature judgment
(Appendix~\ref{app:class2}).

Notably, the best performing LM only scores 72.7\% in the tone-mismatch condition (DS-V4-flash). As distinguishing two characters with different tones is extremely easy for humans, it is surprising to see LMs with $\sim$300B parameters (DS-V4-flash) struggle on this task.

\begin{table*}[t]
\centering
\begin{minipage}{\linewidth}
\small
\textbf{Target character:} 包 (\behavpy{b}{ao}{1})
\quad\textbf{True homophone:} 胞 (\behavpy{b}{ao}{1})\par
\vspace{2pt}
{\small Each row replaces 胞 with a distractor. \behavsame\ = match; \behavdiff\ = mismatch (relative to 包).}\par
\vspace{7pt}
\begingroup
\fontsize{10}{12}\selectfont
\setlength{\tabcolsep}{1.8pt}
\renewcommand{\arraystretch}{1.1}
\begin{tabular*}{\linewidth}{@{\extracolsep{\fill}}l ccc l rrrrrr rr@{}}
\toprule
\textbf{Condition} & \multicolumn{3}{c}{\textbf{Dimensions}} & \textbf{Example} &
\multicolumn{7}{c}{\textbf{Qwen2.5}} & \textbf{DeepSeek}\\
\cmidrule(lr){2-4}\cmidrule(lr){5-5}\cmidrule(lr){6-12}\cmidrule(l){13-13}
& \textcolor{initialfg}{\textbf{Initial}} & \textcolor{finalfg}{\textbf{Final}} & \textcolor{tonefg}{\textbf{Tone}} & Distractor &
0.5B & 1.5B & 3B & 7B & Omni-7B & 14B & 32B & V4\\
\midrule
Disjoint & \behavdiff & \behavdiff & \behavdiff & 地 (\behavpy{d}{i}{4}) & \behavacc{56.3} & \behavacc{66.5} & \behavacc{73.6} & \behavacc{76.8} & \behavacc{88.2} & \behavacc{73.2} & \behavacc{78.7} & \behavacc{98.4}\\
\hdashline
Initial-match & \behavsame & \behavdiff & \behavdiff & 必 (\behavpy{b}{i}{4}) & \behavacc{53.9} & \behavacc{58.7} & \behavacc{63.8} & \behavacc{67.0} & \behavacc{81.8} & \behavacc{65.1} & \behavacc{70.6} & \behavacc{95.1}\\
Final-match & \behavdiff & \behavsame & \behavdiff & 道 (\behavpy{d}{ao}{4}) & \behavacc{53.9} & \behavacc{58.4} & \behavacc{63.7} & \behavacc{68.6} & \behavacc{81.3} & \behavacc{64.5} & \behavacc{69.8} & \behavacc{95.4}\\
Tone-match & \behavdiff & \behavdiff & \behavsame & 低 (\behavpy{d}{i}{1}) & \behavacc{56.9} & \behavacc{64.9} & \behavacc{72.2} & \behavacc{75.9} & \behavacc{88.3} & \behavacc{71.3} & \behavacc{78.0} & \behavacc{97.5}\\
\hdashline
Initial-mismatch & \behavdiff & \behavsame & \behavsame & 刀 (\behavpy{d}{ao}{1}) & \behavacc{53.6} & \behavacc{58.1} & \behavacc{64.0} & \behavacc{67.7} & \behavacc{81.7} & \behavacc{63.1} & \behavacc{67.1} & \behavacc{92.3}\\
Final-mismatch & \behavsame & \behavdiff & \behavsame & 悲 (\behavpy{b}{ei}{1}) & \behavacc{53.9} & \behavacc{59.4} & \behavacc{62.9} & \behavacc{66.7} & \behavacc{80.0} & \behavacc{63.8} & \behavacc{69.1} & \behavacc{90.7}\\
Tone-mismatch & \behavsame & \behavsame & \behavdiff & 抱 (\behavpy{b}{ao}{4}) & \behavacc{51.6} & \behavacc{53.8} & \behavacc{51.1} & \behavacc{54.7} & \behavacc{59.0} & \behavacc{53.3} & \behavacc{55.7} & \behavacc{72.7}\\
\bottomrule
\end{tabular*}
\endgroup\par
\end{minipage}
\caption{
Homophone judgment accuracy (\%) by distractor condition.
Qwen2.5 base models use sentence log-probability comparison;
Omni-7B and DeepSeek-V4 use two-option multiple-choice
questions with one correct option and one distractor.
Chance is 50\% for both protocols.
Characters are illustrative examples.
Pinyin uses numeric tone labels and the initial/final/tone colors of Table~\ref{tab:syllable-decomposition}. 
Condition names identify the unique matching/mismatching dimension.
}
\label{tab:behav-results}
 \vspace{-2mm}
\end{table*}

\subsection{Effect of Phonological Distance and Lexical Frequency (RQ1.2)}
\label{sec:distance}
To address RQ1.2, we examine how phonological distance and lexical
frequency affect performance on the homophone minimal pairs.
Each item is associated with a phonological distance between the
target and distractor, measured using the tone-aware DIMSIM metric
\citep{li2018dimsim}, and a frequency gap $\Delta\mathrm{Zipf}$,
defined as the Zipf frequency of the concatenated target--distractor
string minus that of the target--positive string, both estimated
with the Chinese word lists in \texttt{wordfreq}~\citep{speer2022wordfreq}. We additionally use the segment-only
PanPhon metric \citep{mortensen2016panphon} as a cross-check
(see Appendix~\ref{app:metrics} for definitions).

\begin{figure*}[t]
  \centering
  \includegraphics[width=0.95\linewidth]{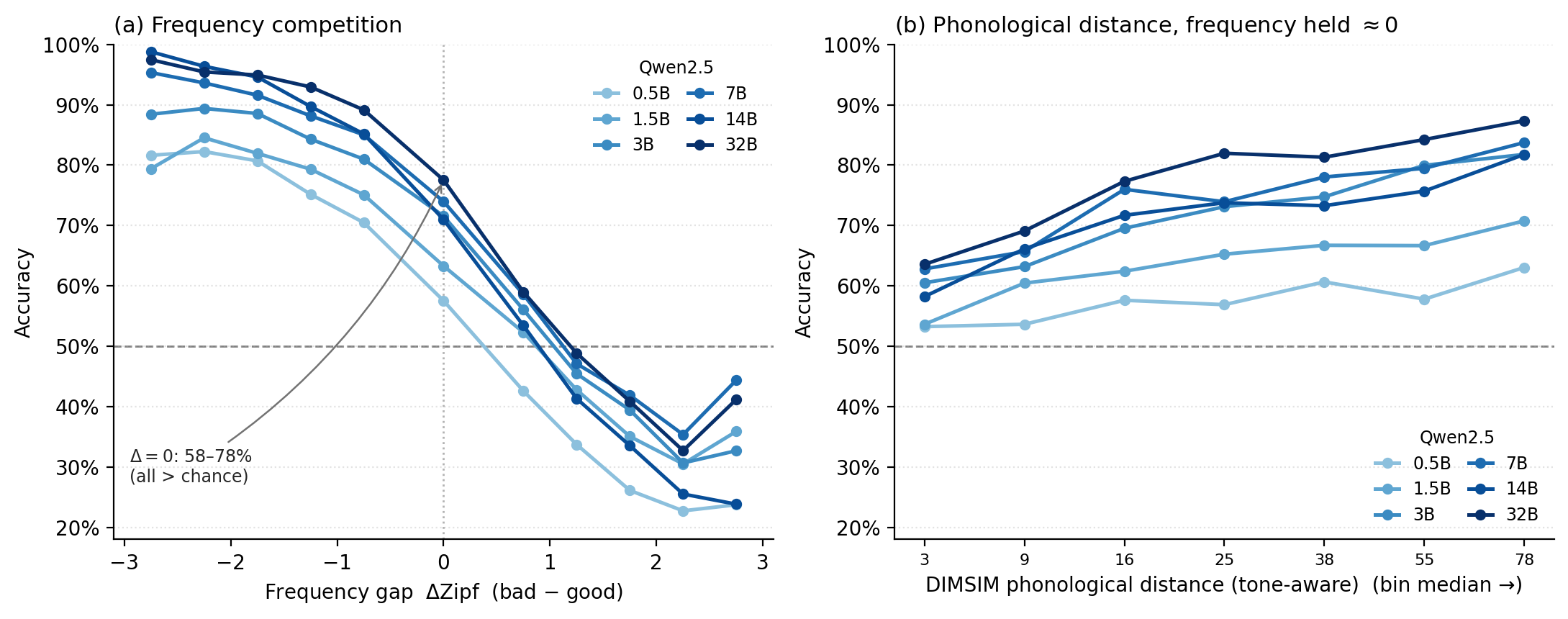}
  \vspace{-5mm}
  \caption{Homophone judgment accuracy vs.\ (a)~frequency gap $\Delta$Zipf and (b)~DIMSIM
  distance ($\Delta$Zipf${\approx}0$). Accuracy generally decreases as the frequency gap grows yet stays above chance at parity; it generally increases with phonological distance. Qwen2.5 base, darker
  $=$ larger.}
  \label{fig:freq-dist}
   \vspace{-2mm}
\end{figure*}

\paragraph{Phonological distance.}
To examine the relationship between phonological distance and
model performance, we focus on items with a frequency gap close
to zero ($\Delta\mathrm{Zipf} \approx 0$).
As shown in Figure~\ref{fig:freq-dist}(b), accuracy increases
with the DIMSIM distance between the target and distractor
across all Qwen2.5 base models. Models find it easier to
reject distractors whose pronunciations are more distinct
from the target, even when the target--distractor and target--positive strings have similar frequencies. This confirms our previous observation that phonological distance independently affects model performance.
The same trend holds with the segment-only PanPhon metric
(Appendix~\ref{app:panphon}).

\paragraph{Lexical frequency.}
Figure~\ref{fig:freq-dist}(a) shows that accuracy decreases
as $\Delta\mathrm{Zipf}$ increases: models find it harder to
identify the correct phonological relation when the distractor is more frequent than the positive candidate.
However, when the frequency gap is close to zero,
all Qwen2.5 base models perform above the 50\% chance level,
with accuracy ranging from 57.5\% for the 0.5B model to
77.5\% for the 32B model. The results suggest that lexical
frequency influences model judgments but does not fully
explain their performance on the monophonic characters.
The same frequency trend is observed in Qwen3 and Qwen3.5
(Appendix~\ref{app:freq-gen}).

Together, these results show that models are sensitive to
phonological differences between characters, while their
judgments are also influenced by lexical frequency.
We next examine how this phonological knowledge is
represented within the models.

\section{Experiment 2: Linear Probing --- Where Phonology Is Decodable}
\label{sec:probing}
\subsection{Setup}
\label{sec:probing-setup}

We use linear probing to examine where information about
initials, finals, and tones can be recovered from model
representations, and how this varies with prompt context.

We use 2,300 monophonic characters drawn from the
3,500-character pool introduced in
Section~\ref{sec:benchmark}.
For each character, we extract its representation at
the input embedding layer and after each transformer
layer, and train separate linear classifiers to predict
its initial, final, and tone.

To examine the role of context, we present each character
under four prompt conditions: bare (the character alone),
phonetic (这个字的读音：; ``the pronunciation of this
character''), neutral (这个字是：; ``this character is''),
and semantic (这个字的意思：; ``the meaning of this
character'').
In each condition, the character appears at the end
of the input, and we extract its representation at
that position.
We train and evaluate probes separately for each
layer and prompt condition.

We evaluate prediction performance on held-out characters
using five-fold cross-validation and report macro-$F_1$,
which gives equal weight to each phonological class.
We also compare performance against probes trained
on shuffled labels, reporting the difference as
\emph{selectivity} \citep{hewitt2019control}.
Character filtering, representation extraction, and
probe training details are provided in
Appendix~\ref{app:probing-setup}.

\subsection{Results}
\label{sec:exp2:res}

\begin{figure*}[t]
  \centering
  \includegraphics[width=0.98\textwidth]{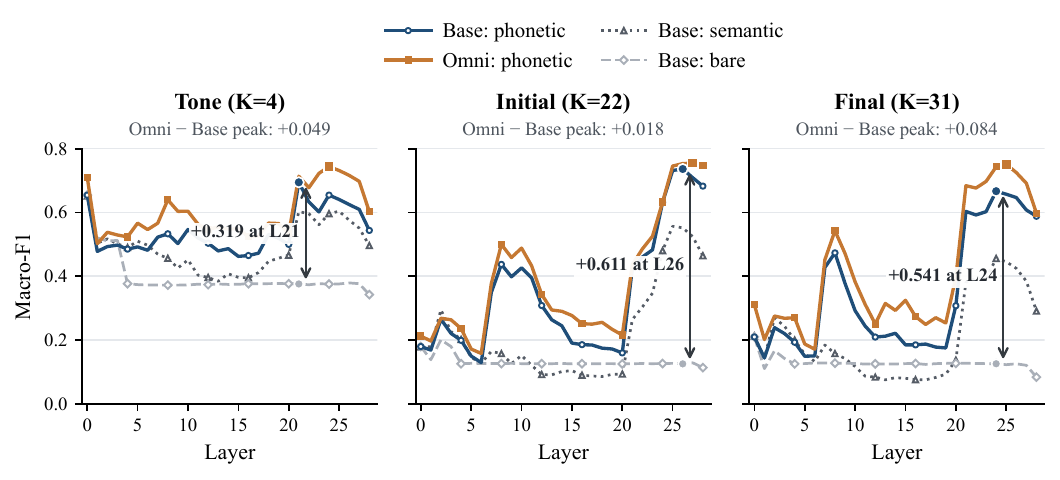}
  \vspace{-5mm}
\caption{
\textbf{Phonological decodability across layers.}
Macro-$F_1$ of linear probes for Qwen2.5-7B-base
and Qwen2.5-Omni-7B condition.
$K$ denotes the number of classes.
Arrows show the phonetic--bare difference for Base
at its phonetic-prompt peak layer.
Values below panel titles show the difference
between Omni and Base at their respective peaks.
Tone shows the smallest phonetic--bare gain.
The full four-prompt comparison, including the
neutral condition, appears in
Appendix~\ref{app:fourprompt}.
}
\label{fig:probing-rq123-revised}
 \vspace{-2mm}
\end{figure*}

The overall results are presented in Figure~\ref{fig:probing-rq123-revised}.
We make the following observations. 

\paragraph{All three phonological dimensions are linearly decodable from the model's hidden states under the phonetic prompt.} 
Tone shows the smallest margin over its shuffled-label control, partly because chance is higher with only four classes: its selectivity is $0.482$, compared with $0.708$ for initials and $0.708$ for finals, despite their much larger label inventories. Our results further show that tone differs from segmental features.

\paragraph{Phonetic prompts improve segmental decodability more.}
In Qwen2.5-7B-base, phonetic prompts yield higher
decodability than semantic prompts or bare characters,
with larger gains for initials and finals than for tone
(Figure~\ref{fig:probing-rq123-revised}).
Tone remains comparatively decodable even without
surrounding context.

\paragraph{Initial and final decodability improves more
with depth than tone decodability.}
Tone is already strongly decodable from the character
embedding and improves only modestly in subsequent layers
(Figure~\ref{fig:probing-rq123-revised}).
Initial and final decodability starts substantially lower
but rises sharply across layers, peaking deeper in the model.
Despite its strong decodability at the embedding layer,
tone remains the weakest dimension in the behavioral
benchmark (Experiment~1).
This contrast suggests that models encode tone information
but struggle to use it effectively when making explicit
phonological judgments. One possibility is that recognizing an individual
character's tone is easier than comparing tones
across characters.

\paragraph{The multimodal model improves decodability
but retains a weakness in tone judgments.}
Using the same probing pipeline, we compare
Qwen2.5-Omni-7B with Qwen2.5-7B-base to examine
phonological representations in multimodal and
text-only models.
Omni achieves higher macro-$F_1$ at both \layer{0}
and the peak layer for all three dimensions,
with the largest improvement for final and the
smallest for initial
(Figure~\ref{fig:probing-rq123-revised}).
The overall layerwise pattern remains similar:
tone is already strongly decodable at the embedding
layer, whereas initial and final decodability
improves more with depth.
However, tone remains Omni's weakest dimension
in the behavioral benchmark
(Experiment~1; Table~\ref{tab:behav-results}).
Together, these findings suggest that richer
phonological representations in a multimodal model
do not necessarily translate into reliable
tone judgments.

\section{Experiment 3: Activation Steering --- Steering Phonological Outputs}
\label{sec:steering}

Experiment~2 shows that phonological information
is decodable from model representations.
We next test whether interventions on these
representations can shift outputs toward a
target phonological class.

\subsection{Setup}
\label{sec:steering-setup}
In mechanistic interpretability, activation steering modifies a model's internal representations during inference to influence its output \citep{rimsky2024caa}. We test whether steering can shift output probabilities
from characters in one phonological class to those in another. For example, given a prompt asking for a character with the
same initial as 吉~(\hlinitial{j}\hlfinal{i}\hltone{2}),
we aim to increase the probability of characters with initial
\hlinitial{l}. We apply the same approach to contrasts in
initial, final, and tone.

For each contrast, we construct a steering vector by averaging
representation differences between character pairs that match
on the other two phonological dimensions. We add this vector
to the source character's representation during the forward
pass, keeping the prompt unchanged. By varying the intervention
strength and comparing the next-token probabilities of the
source and target classes, we measure how strongly the output
responds to each phonological contrast.
\orange{
For technical details of the experiment, see Appendix~\ref{app:steering:setup}.
}

\subsection{Results}
\label{sec:steering-graded}
\label{sec:steering-emb}
\begin{figure}[t]
  \centering
  \includegraphics[width=\linewidth]{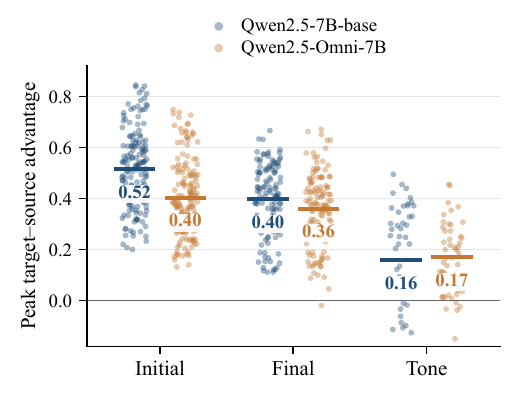}
  \vspace{-5mm}
  \caption{
\textbf{Peak target--source advantage by dimension.}
Each point shows $R_{\mathrm{peak}}$ for one
transition at one intervention site
(Emb/L8/L13/L20); bars indicate means.
Zero denotes equal target and source probabilities.
}
  \label{fig:dist}
\end{figure}

\paragraph{Tone Shows the Smallest Steering Effect.} Tone yields the smallest steering effect, with a mean peak target--source advantage of $0.16$,
compared with $0.52$ for initials and $0.40$ for finals
(Figure~\ref{fig:dist}).
This ordering holds at all sites
(Table~\ref{tab:steer-layer}). This weaker steering response also agrees with its poorer
behavioral performance (\S\ref{sec:behavioral}), despite being highly
decodable from the character embedding (\S\ref{sec:probing}).

Omni's higher probing scores do not make it any easier to steer (Figure~\ref{fig:dist}). It follows a similar pattern to the base model, where tone is the least steerable dimension. \orange{Initials and finals are even harder to steer for the omni model. }

\begin{table}[t]
\centering
\small
\setlength{\tabcolsep}{3pt}
\resizebox{\columnwidth}{!}{%
\begin{tabular}{lcccccc}
\toprule
Site & Initial & Final & Tone & Switch & Flip cost & Collapse \\
     &         &       &      & (init.) & (init.) &          \\
\midrule
Emb & 0.575 & 0.452 & 0.164 & 1.00 & 0.64 & 0.03 \\
L8  & 0.535 & 0.405 & 0.136 & 1.00 & 0.40 & 0.15 \\
L13 & 0.514 & 0.398 & 0.171 & 1.00 & 0.36 & 0.20 \\
L20 & 0.461 & 0.377 & 0.153 & 1.00 & 0.31 & 0.30 \\
\bottomrule
\end{tabular}%
}
\caption{
Steering Qwen2.5-7B-base at the four intervention sites
summarized in the main text.
Initial, Final, and Tone report mean peak steering
advantage ($R_{\mathrm{peak}}$) across transitions
in each dimension.
\emph{Switch}: fraction of initial transitions that flip.
\emph{Flip cost}: relative perturbation
$\lVert\alpha v_{\mathrm{dir}}\rVert/\lVert h\rVert$
at the flip point.
\emph{Collapse}: rate of response decline after the peak
(Appendix~\ref{app:steer-alpha}).
}
\label{tab:steer-layer}
 \vspace{-2mm}
\end{table}

\begin{figure}[t]
  \centering
  \includegraphics[width=\linewidth]{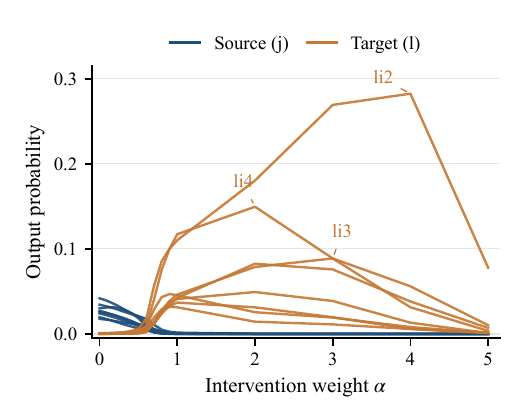}
 \vspace{-5mm}
\caption{
Character output probabilities under initial steering
from \textit{j} to \textit{l} at L8 in Qwen2.5-7B-base.
Each curve represents one character:
eight source characters (blue) and eight target
characters (orange).
Selected target curves are labeled with Pinyin.
}
\label{fig:exemplar}
\end{figure}

\paragraph{Steering Shifts Character Probabilities.}
\label{sec:steering-transfer}
Figure~\ref{fig:exemplar} illustrates steering from \hlinitial{j}
to \hlinitial{l} at \layer{8} in Qwen2.5-7B-base.
The prompt asks for a character with the same initial as
吉~(\hlinitial{j}\hlfinal{i}\hltone{2}).
After intervention, the probabilities of the plotted
source-initial characters decrease, while those of multiple
target-initial characters increase.
Additional initial and final examples are reported in
Appendix~\ref{app:steer-trans}.

\paragraph{Steering Extends to Held-Out Characters.}
We next examine target-class characters that were not used
to construct the steering direction. For six selected initial
transitions at \layer{8}, $76$--$85\%$ of these characters
increase in output probability after intervention
(Appendix~\ref{app:heldout}). The effects therefore extend
beyond the characters used to construct each direction.

\section{Discussion and Conclusion}
\label{sec:discussion}

Our findings reveal a systematic dissociation between
segmental and tonal knowledge in Chinese LMs, and,
more broadly, between the encoding and functional
accessibility of phonological information. Across models,
initials and finals are more reliably discriminated
than lexical tones. Although character frequency
influences performance, phonological distance also
predicts accuracy, indicating that models possess
phonological knowledge beyond character familiarity.

Crucially, the three experiments reveal different
aspects of this knowledge. Probing shows that tone
is highly decodable even from character embeddings,
despite near-chance performance on tone-only contrasts.
In contrast, segmental information becomes increasingly
decodable in deeper layers and is more responsive
to activation steering. This asymmetry is compatible
with the distinction between tonal and segmental
structure in autosegmental phonology
\citep{goldsmith1976autosegmental}, although our results
do not establish that LMs encode such representations.
Notably, manipulating character embeddings alone
can systematically shift output probabilities toward
target initial and final classes, extending previous
findings on phonological steering in alphabetic languages
\citep{mclaughlin2025rhyme}. These results demonstrate
that linear decodability does not guarantee effective
use in linguistic judgments or generation.

The comparison between text-only and audio-trained
models reinforces this distinction. Qwen2.5-Omni
exhibits stronger phonological decodability than
Qwen2.5-7B-base, including for tone, without a
corresponding advantage in steering. Thus,
multimodal training is associated with richer decodable
phonological information, but not necessarily greater
controllability under the tested protocol.

Overall, our study demonstrates that phonological
knowledge in Chinese LMs is multidimensional:
what models encode, what they can judge, and what
can be manipulated are not equivalent.
Our findings motivate studying how encoded
phonological information supports pronunciation
judgments and context-dependent reading selection.

\section*{Limitations}
\addcontentsline{toc}{section}{Limitations}
The behavioral hierarchy is measured across seventeen models, but the representational and
intervention analyses center on the Qwen2.5 base family, with steering tested at five representative depths of Qwen2.5-7B-base and Qwen2.5-Omni-7B. Steering is read through class probability mass under
elicitation prompts, not free generation, and the held-out analysis shows class-level
spread rather than explaining why individual characters rise. The tone results concern
isolated-character lexical tones---not neutral tone, tone sandhi, or the contextual
reading selection that Task~3 shows is a distinct challenge. The Omni comparison probes
the text backbone only, so it attributes changes to multimodal training as a whole rather
than to speech input specifically; and because the models' training data is unknown, we
characterize representation and use rather than origin.

\section*{Ethics Statement}
\addcontentsline{toc}{section}{Ethics Statement}
This work evaluates publicly available language models on isolated Chinese characters and
their pronunciations, using standard public character and frequency resources; it involves
no human subjects and no personal or sensitive data, and steering modifies activations
only transiently at inference time without altering model weights. We see no foreseeable
harm.

\section*{Reproducibility and Data Availability}
\addcontentsline{toc}{section}{Reproducibility and Data Availability}
All experiments draw on one character pool (the $3{,}500$ most common characters plus the $3{,}500$--$7{,}000$ frequency band) and one IPA-based phonological labeling (pypinyin for
pinyin, Dragonmapper for IPA); the probe and steering configurations are fully specified
in \S\ref{sec:probing-setup} and \S\ref{sec:steering-setup}, with random seeds fixed
($42$) throughout. The benchmark items, IPA phonological labels, the $82$ steering
transitions, probing and steering code, and per-layer result tables will be released in a
public repository upon publication.

\section*{Use of AI Assistance}
\addcontentsline{toc}{section}{Use of AI Assistance}
An AI assistant helped draft and organize the manuscript prose from the authors'
experimental records and outline; all design, data, analysis, numerical results, and
citations are the authors' own and were verified by them.

\bibliography{references}

\appendix

\section{IPA-Based Initial/Final Split}
\label{app:ipa-split}
Pinyin is obtained with pypinyin (a handful of characters it mislabeled were corrected by
hand) and mapped to IPA with the Dragonmapper library, whose syllable inventory supplies
each syllable's IPA. We split a syllable into an initial and a final by reverse-deriving
every initial's IPA prefix from that inventory---the longest common IPA prefix shared by
all syllables with the same pinyin initial---and taking the final as the remainder (with
tone-value diacritics stripped); tone is kept as an integer $1$--$4$.

\section{Explicit Phonological Feature Judgments}
\label{app:class2}
\begin{figure*}[t]
  \centering
  \includegraphics[width=0.99\linewidth]{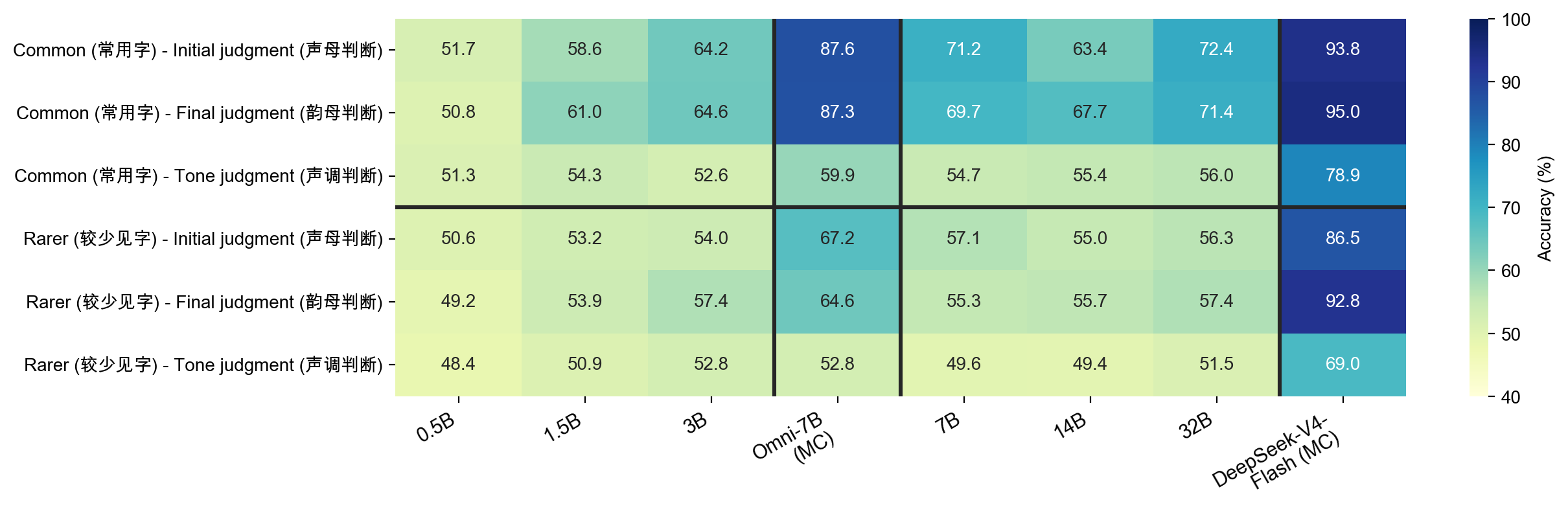}
  \caption{Feature judgment accuracy by model and character-frequency band.
  Unboxed: Qwen2.5 base (0.5B--32B, zero-shot log-probability); boxed: Omni-7B and
  DeepSeek-V4-Flash (multiple-choice). Initial and final judgments are generally stronger than
  tone judgments, and the rarer-character subset is harder. Models as in Table~\ref{tab:behav-results}; chance $=50\%$.}
  \label{fig:class2}
\end{figure*}

Figure~\ref{fig:class2} reports feature judgment results for the same Qwen2.5 base checkpoints
and multiple-choice reference models as Table~\ref{tab:behav-results}, split by
character-frequency band. It supports the converging behavioral result summarized in
\S\ref{sec:behavioral}: a second elicitation yields the same ordering, with segmental
judgments generally stronger than tone judgments,
while the rarer-character subset is generally harder.

\section{Polyphone Disambiguation}
\label{app:class3}

\begin{figure*}[t]
  \centering
  \includegraphics[width=0.92\linewidth]{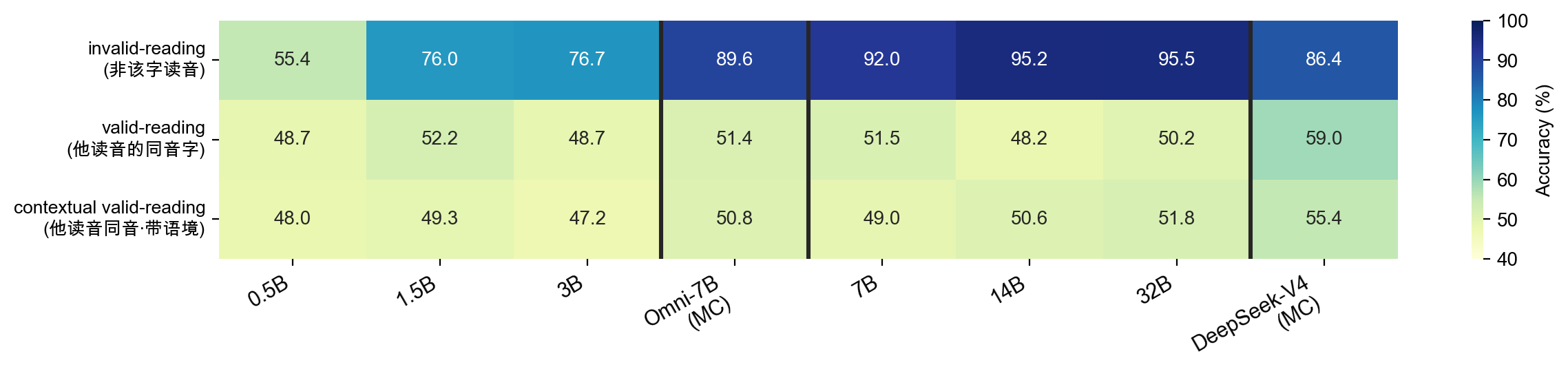}
  \caption{Polyphone disambiguation accuracy by subtype and model. Invalid-reading rejection
  improves with scale, but valid-reading and contextual disambiguation stay near chance
  for all models. Models as in Table~\ref{tab:behav-results}; chance $=50\%$.}
  \label{fig:class3}
\end{figure*}

\paragraph{Model performance on polyphonic characters.} Apart from monophonic characters, we also examine polyphonic characters to see if models can correctly identify their context-appropriate pronunciations.
Models improve with scale at rejecting invalid readings,
but remain near chance when the distractor corresponds to
another valid reading of the same character, including
for DeepSeek-V4 (Figure~\ref{fig:class3}).
These results suggest that selecting a pronunciation in
context remains difficult even when models can distinguish
phonological contrasts in monophonic characters.
The three polyphone subtypes split sharply (Figure~\ref{fig:class3}): rejecting an
\emph{invalid reading} is learnable and improves with scale (${\sim}95\%$ for the largest
models), but choosing the correct reading over a homophone of the character's \emph{other}
valid reading---or its contextual variant---stays near chance for \emph{every} model, including
DeepSeek-V4. The bottleneck is thus \emph{context-conditioned selection} among readings, \orange{not
knowing which readings a character has---a disambiguation problem} distinct from the
segmental/tonal representation question we study here. We therefore restrict the
representational analyses (\S\ref{sec:probing}--\S\ref{sec:steering}) to monophonic characters
and leave contextual reading selection to future work.

Figure~\ref{fig:class3} breaks polyphone disambiguation accuracy down by subtype and model,
quantifying the context-conditioned selection bottleneck described above.

\begin{table*}[t]
  \centering
  \small
  \renewcommand{\arraystretch}{1.15}
  \setlength{\tabcolsep}{5pt}
  \begin{tabular}{@{}>{\raggedright\arraybackslash}l l l >{\raggedright\arraybackslash}p{4.2cm}@{}}
  \toprule
  \textbf{Task \& Prompt} & \textbf{Positive} & \textbf{Distractor} & \textbf{Condition} \\
  \midrule
  \textbf{Polyphone Disambiguation} & 涨 (zhǎng) & 弃 (qì) & invalid-reading \\
  “C”的“A”和“B”读音相同 & 涨 (zhǎng) & 常 (cháng) & valid-reading \\
  (``A'' read in the word C sounds the same as B) & 涨 (zhǎng) & \textit{经常}的\textit{常}~(cháng) & contextual valid-reading \\
  \bottomrule
  \end{tabular}
  \caption{PinyinBench Polyphone Disambiguation, built from the target character \textit{长}~(zhǎng) read
  in the word \textit{生长}. It is presented separately because it probes context-conditioned
  reading \emph{selection} rather than a single initial/final/tone contrast, and is therefore
  analyzed apart from the representational experiments.}
  \label{tab:pinyinbench-polyphone}
\end{table*}

\section{Frequency Curves for Qwen3 and Qwen3.5}
\label{app:freq-gen}
\begin{figure*}[t]
  \centering
  \includegraphics[width=\linewidth]{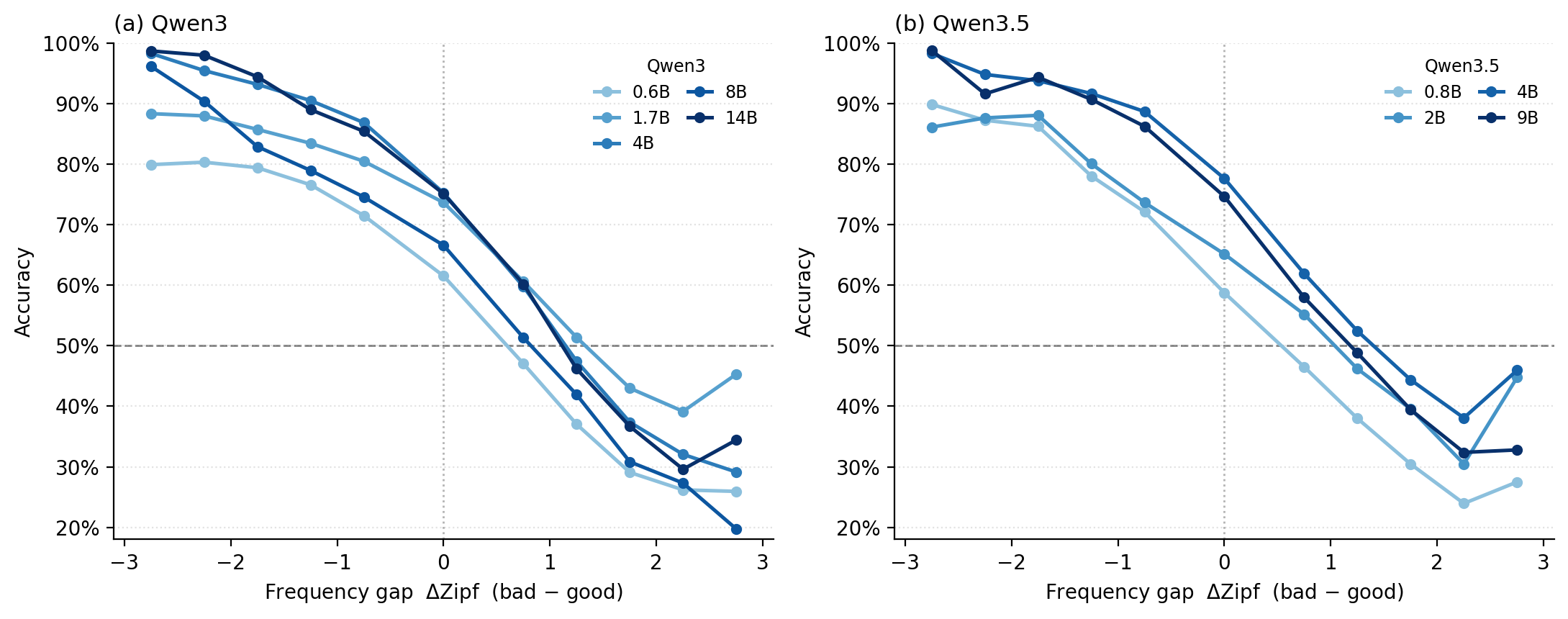}
  \caption{Accuracy vs.\ frequency gap for (a)~Qwen3 and (b)~Qwen3.5 base models,
  replicating Figure~\ref{fig:freq-dist}a: accuracy declines $10$--$14$~pp per unit
  $\Delta$Zipf ($R^2>0.92$) for every model, and every model stays above chance in the
  near-zero bin ($59$--$78\%$).}
  \label{fig:freq-gen}
\end{figure*}

Figure~\ref{fig:freq-gen} replicates the frequency analysis of \S\ref{sec:distance} on the
two later model generations.

\section{Phonological Distance Metrics: Formal Definitions}
\label{app:metrics}
Both metrics measure the distance between the pronunciation of the \emph{target} character $B$
and that of the \emph{distractor} character $N$; a larger distance means a more obviously
different distractor character. (In the released data these are the fields \texttt{good1\_pinyin} and
\texttt{bad2\_pinyin}.)

\paragraph{DIMSIM (tone-aware).}
DIMSIM \citep{li2018dimsim} models a pinyin syllable as three independent components---initial
$p^{I}$, final $p^{F}$, tone $p^{T}$---and defines the distance between two characters as the
Manhattan sum of three component distances,
\begin{equation}
\resizebox{\columnwidth}{!}{$\displaystyle d_{\textsc{DimSim}}(B,N) = S_p\!\left(p^{I}_B,p^{I}_N\right) + S_p\!\left(p^{F}_B,p^{F}_N\right) + S_T\!\left(p^{T}_B,p^{T}_N\right),$}
\label{eq:dimsim}
\end{equation}
where $S_p$
is the Euclidean distance between learned low-dimensional encodings of initials (or finals)
and $S_T$ is a function of the tone numbers. The released implementation realizes
$S_p$ with a fixed two-dimensional coordinate table and computes a per-syllable score
$\min\!\left(d_{I}+d_{F},\,d_{\mathrm{hc}}\right) + \tfrac{1}{10}\lvert\Delta\text{tone}\rvert$
(with $d_{\mathrm{hc}}$ a hard-coded near-homophone shortcut), which it then multiplies by a
ratio that drives the tone contribution to ${\sim}0$. We make three corrections so tone is not
washed out: (i)~drop the suppressing ratio; (ii)~split the zero-onset class into a true
zero-onset (bare-vowel) coordinate and separate glide-onset coordinates for the
\textit{y}-, \textit{w}-, and \textit{yu}-initial syllables, removing a coordinate-blowup
that sent some distances to ${\sim}10^{5}$; and (iii)~scale tone by $3$ so it is commensurate with the segmental
coordinate distances. The per-character distance we use is therefore
\begin{equation}
d_{\textsc{DimSim}}(B,N) \;=\; \min\!\left(d_{I}+d_{F},\,d_{\mathrm{hc}}\right) \;+\; 3\,\lvert\Delta\text{tone}\rvert .
\label{eq:dimsim-ours}
\end{equation}
For example $d_{\textsc{DimSim}}(\text{an4},\text{an2})=0+0+3{\cdot}2=6$ (tone only), while
$d_{\textsc{DimSim}}(\text{an4},\text{dao3})=33.1+4.1+3.0=40.2$ (all three components differ).

\paragraph{PanPhon (segmental).}
PanPhon \citep{mortensen2016panphon} maps each IPA segment to a numeric articulatory feature
vector $v\in\{-1,0,+1\}^{F}$ ($F{\approx}24$ features). The cost of substituting segment $a$ for
$b$ is the weighted $L_1$ difference of their feature vectors, $\sum_{f} w_f\,\lvert v^{a}_f -
v^{b}_f\rvert$, and an insertion or deletion costs $\sum_{f} w_f$; the distance between two
syllables is the minimum-cost feature edit (Wagner--Fischer) alignment of their IPA segment
sequences. Because the pinyin-to-IPA step strips tone diacritics, $d_{\textsc{PanPhon}}$ is
purely segmental and tone-blind---hence its role as a cross-check rather than a primary metric.

\section{Phonological Distance under the Segmental PanPhon Metric}
\label{app:panphon}
\begin{figure*}[t]
  \centering
  \includegraphics[width=\linewidth]{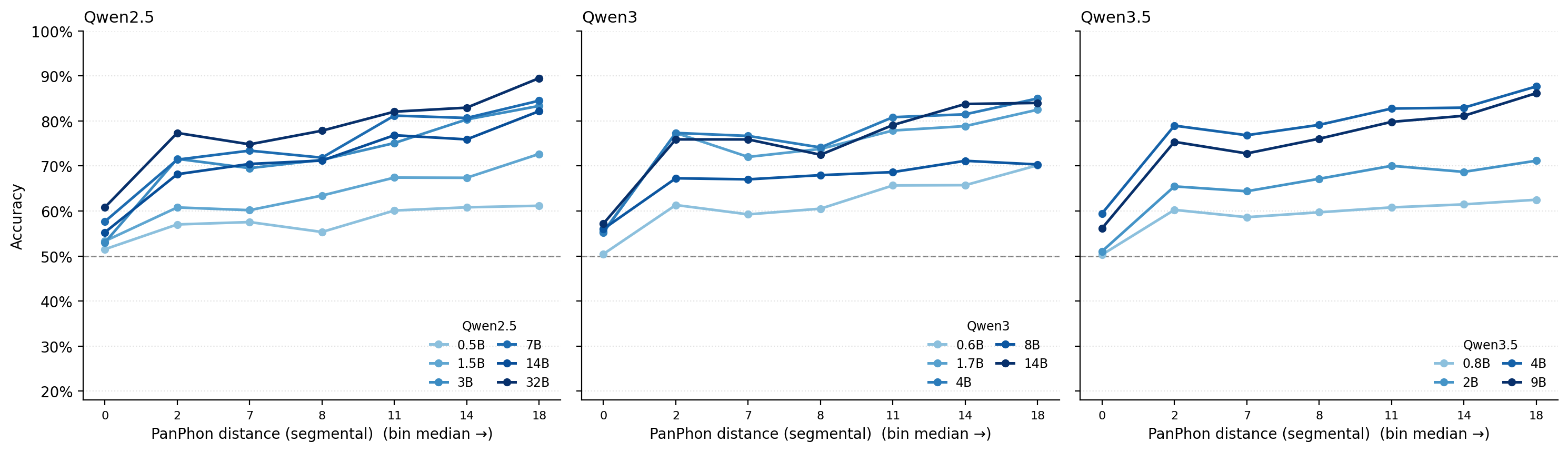}
  \caption{Accuracy vs.\ PanPhon distance ($\Delta$Zipf${\approx}0$) for Qwen2.5, Qwen3,
  and Qwen3.5 base models. The monotonic rise mirrors DIMSIM (Figure~\ref{fig:freq-dist}b)
  across all generations; because PanPhon is tone-blind, the effect's survival shows the
  distance--difficulty relationship is carried largely by segmental structure.}
  \label{fig:panphon}
\end{figure*}

Figure~\ref{fig:panphon} repeats the distance analysis of \S\ref{sec:distance} under the
segment-only PanPhon metric (defined in Appendix~\ref{app:metrics}).

\section{Joint Logistic Regression: Coefficients and Effect Sizes}
\label{app:logit}
This appendix backs the separability claim of \S\ref{sec:distance}. For each model we fit
\begin{equation}
\operatorname{logit}\Pr(\text{correct}) = \beta_0 + \beta_1\,\tilde z(\Delta\mathrm{Zipf}) +
\beta_2\,\tilde z(d_{\textsc{DimSim}})
\label{eq:logit}
\end{equation}
on all $23{,}742$ homophone judgment items, with both predictors
standardized (so $\beta$ is an effect \emph{per standard deviation}) and standard errors
clustered on the target pronunciation (the field \texttt{good1\_pinyin}; the $2{,}690$ distinct
target characters collapse to only ${\sim}709$ pinyin pronunciations, each reused ${\sim}33$ times, so
ordinary SEs are anti-conservative; clustering at this coarser pronunciation level---more
conservative than clustering on the character---inflates them ${\sim}1.7$--$3.1\times$). The two predictors are essentially
orthogonal (Pearson $r=-0.015$, VIF ${\approx}1$), so neither is a proxy for the other.
Table~\ref{tab:logit} reports, per predictor, the standardized coefficient and the average
marginal effect on accuracy (percentage points per SD); the last
column is the distance coefficient after adding condition (\texttt{task}) fixed effects---a
stricter, within-condition test. Every frequency coefficient is significant at
$p<10^{-3}$; every distance coefficient is positive and significant, with Qwen3.5-0.8B at
$p=0.015$ and the rest at $p<10^{-3}$.

\begin{table}[h]
\centering
\small
\setlength{\tabcolsep}{4pt}
\caption{Joint logistic regression of homophone judgment correctness on standardized frequency gap and
DimSim distance (cluster-robust on the target pronunciation). AME is the average marginal
effect on accuracy in percentage points per SD; the odds ratio per $+1$~SD is $e^{\beta}$.
All $\beta_{\mathrm{freq}}$ terms are significant at $p<10^{-3}$; all $\beta_{\mathrm{dist}}$
terms are positive and significant, with Qwen3.5-0.8B at $p<0.05$ and the rest at $p<10^{-3}$.
The final column, $\beta_{\mathrm{dist}}^{\,\mathrm{cond}}$, is the distance coefficient with
\texttt{task} fixed effects added (within-condition test); $^{***}p<10^{-3}$,
$^{**}p<10^{-2}$, $^{*}p<0.05$, $^{\dagger}p<0.1$, ns otherwise.}
\label{tab:logit}
\begin{tabular}{l cc cc c}
\toprule
 & \multicolumn{2}{c}{Freq.\ gap $\Delta$Zipf} & \multicolumn{2}{c}{Dist.\ (DimSim)} & \\
\cmidrule(lr){2-3}\cmidrule(lr){4-5}
Model & $\beta$/SD & AME & $\beta$/SD & AME & $\beta_{\mathrm{dist}}^{\,\mathrm{cond}}$ \\
\midrule
\multicolumn{6}{l}{\emph{Qwen2.5}}\\
\quad 0.5B & -1.04 & -20.5 & 0.08 & +1.6 & 0.05* \\
\quad 1.5B & -0.89 & -18.0 & 0.16 & +3.2 & 0.05$^{\dagger}$ \\
\quad 3B & -1.11 & -19.9 & 0.28 & +5.1 & 0.08** \\
\quad 7B & -1.20 & -19.9 & 0.31 & +5.2 & 0.11*** \\
\quad 14B & -1.58 & -23.9 & 0.31 & +4.7 & 0.09** \\
\quad 32B & -1.44 & -21.6 & 0.37 & +5.5 & 0.09* \\
\multicolumn{6}{l}{\emph{Qwen3}}\\
\quad 0.6B & -0.94 & -19.1 & 0.13 & +2.7 & 0.08** \\
\quad 1.7B & -0.90 & -16.8 & 0.25 & +4.6 & 0.10*** \\
\quad 4B & -1.45 & -22.3 & 0.33 & +5.1 & 0.12*** \\
\quad 8B & -1.18 & -21.4 & 0.20 & +3.7 & 0.13*** \\
\quad 14B & -1.48 & -22.5 & 0.31 & +4.7 & 0.11** \\
\multicolumn{6}{l}{\emph{Qwen3.5}}\\
\quad 0.8B & -1.11 & -21.1 & 0.10 & +2.0 & 0.04~ns \\
\quad 2B & -0.93 & -18.1 & 0.14 & +2.8 & 0.01~ns \\
\quad 4B & -1.30 & -20.0 & 0.35 & +5.4 & 0.05~ns \\
\quad 9B & -1.37 & -21.6 & 0.35 & +5.5 & 0.08** \\
\bottomrule
\end{tabular}
\end{table}

\section{Controlled-Position Probing across Four Templates}
\label{app:fourprompt}

\begin{figure*}[t]
  \centering
  \includegraphics[width=0.9\linewidth]{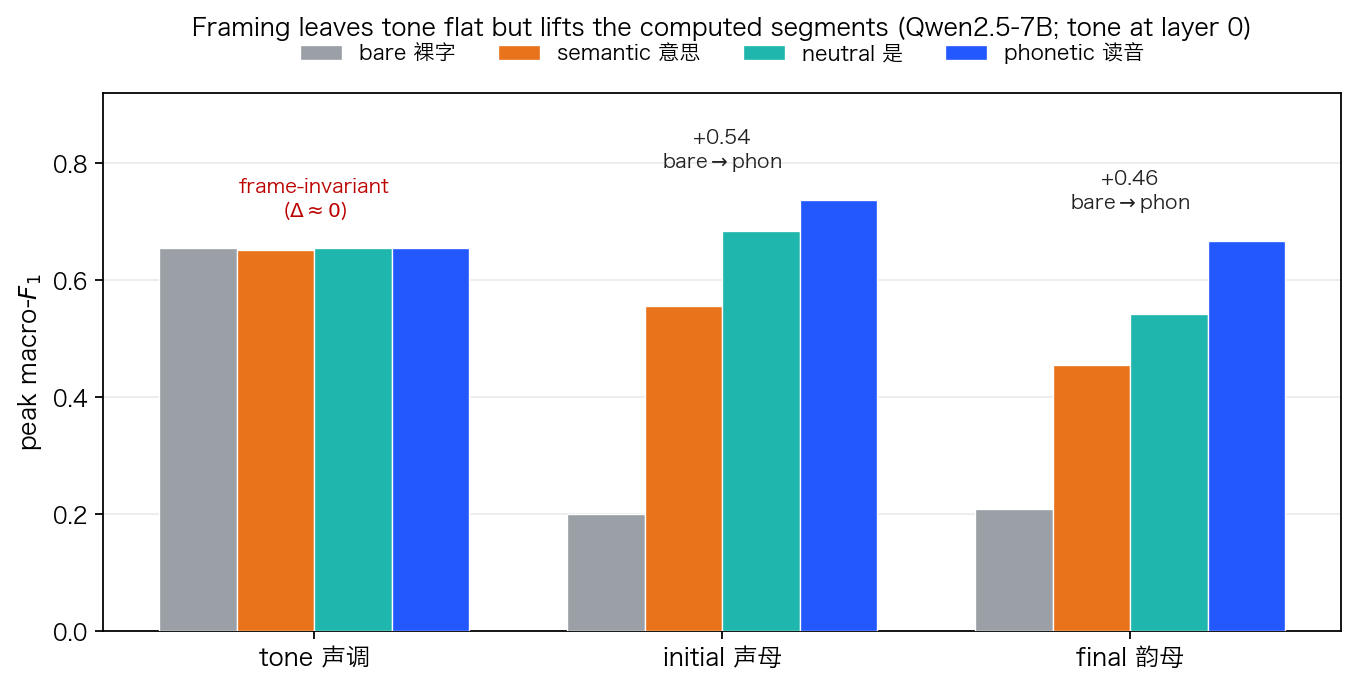}
  \caption{Template comparison for Qwen2.5-7B (controlled position):
tone macro-$F_1$ at \layer{0} and peak macro-$F_1$ over layers for initials and finals.
Tone scores coincide at \layer{0}, where the character embedding is shared across prompts;
contextual prompts improve segmental decodability.}
  \label{fig:fourbars}
\end{figure*}

Figure~\ref{fig:fourbars} compares tone macro-$F_1$ at \layer{0}
and peak segmental macro-$F_1$ across the four templates,
complementing the framing analysis of \S\ref{sec:exp2:res}.

\section{Controlled-Position Probing for the Smaller Qwen2.5 Models}
\label{app:probing-small}

\begin{figure*}[t]
  \centering
  \includegraphics[width=\linewidth]{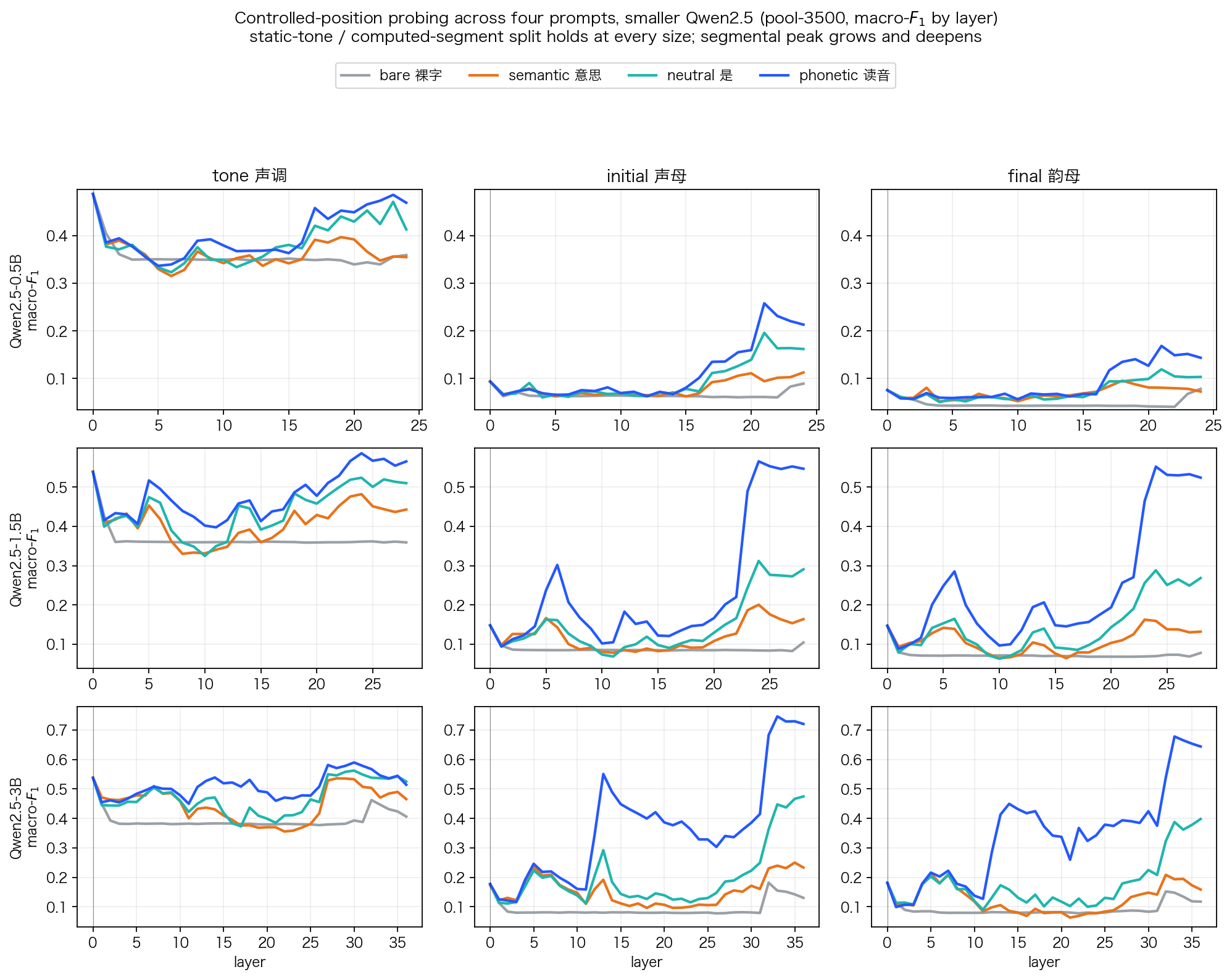}
  \caption{Controlled-position probing for Qwen2.5-0.5B, 1.5B, and 3B.
Tone is already decodable at \layer{0} and changes comparatively modestly with depth,
whereas segmental decodability rises substantially in deeper layers.
Across 0.5B--7B, peak macro-$F_1$ increases from $0.258$ to $0.736$ for initials,
from $0.169$ to $0.667$ for finals, and from $0.487$ to $0.695$ for tone.
The initial peak moves from \layer{21} at 0.5B to \layer{33} at 3B.
Selectivity rises in step with macro-$F_1$, so the segmental gains are not
explained by the probe fitting arbitrary labels.}
  \label{fig:probing-small}
\end{figure*}

Figure~\ref{fig:probing-small} repeats the controlled-position probing
for the smaller Qwen2.5 base models.

\section{Held-Out Test: Steering a Class, Not Memorized Characters}
\label{app:heldout}
Each of the six initial steering vectors in Table~\ref{tab:heldout} is constructed from only $16$--$29$ characters, while the target phonological class contains $397$--$1{,}020$ single-token characters;
Table~\ref{tab:heldout} reports how many of the \emph{un-constructed} same-class characters
rise after steering at \layer{8}.

\begin{table}[h]
\centering
\small
\caption{Held-out class test at \layer{8}: share of same-class characters \emph{not} used
to build the steering vector whose probability rises under steering. Over $96\%$ of all
rising characters were never in the construction set, so steering moves the whole
phonological class rather than a memorized character list.}
\label{tab:heldout}
\begin{tabular}{lc}
\toprule
Transition & Held-out risers \\
\midrule
\ipa{tɕ}$\to$\ipa{l}            & $85\%$ \\
\ipa{l}$\to$\ipa{p}             & $84\%$ \\
\ipa{p}$\to\varnothing$         & $82\%$ \\
\ipa{pʰ}$\to$\ipa{l}            & $81\%$ \\
\ipa{ɕ}$\to\varnothing$         & $80\%$ \\
\ipa{m}$\to\varnothing$         & $76\%$ \\
\bottomrule
\end{tabular}
\end{table}

\section{Prompt-Cue Control}
\label{app:cue}
Each elicitation prompt names the source character (e.g.\ ``\,{"吉"的同声母字}\,''), so the
steering direction is injected at the source-character position. To confirm that the effect
depends on this source-character cue rather than on the surface wording of the prompt, we
compare prompts that carry the cue---including a more explicit tone variant
(``\,{"X"是第\,$N$\,声，和它同声调的字}\,'')---against an otherwise-matched prompt that omits
the source character (so the direction is injected at the final position), holding
transitions, sites, and the $\alpha$ grid fixed (Table~\ref{tab:cue}).

\begin{table}[h]
\centering
\footnotesize
\setlength{\tabcolsep}{4pt}
\caption{Prompt-cue control on Qwen2.5-7B-base: peak target$-$source advantage, mean over
Emb/\layer{8}/\layer{13}/\layer{20}, and the resulting tone class-mass (four-class baseline
$0.25$). The cue is necessary for all three dimensions; given the cue, tone crosses the
baseline but remains the smallest effect.}
\label{tab:cue}
\begin{tabular}{lcccc}
\toprule
Prompt & \initial{} & \final{} & \tone{} & \tone{} mass \\
\midrule
cued (parallel)        & $+0.52$ & $+0.41$ & $+0.16$ & $0.35$ \\
cued (explicit tone)   & $+0.52$ & $+0.41$ & $+0.25$ & $0.41$ \\
no cue                 & $-0.13$ & $-0.07$ & $-0.03$ & $0.17$ \\
\bottomrule
\end{tabular}
\end{table}

\section{Setup of Exp 3: Steering}
\label{app:steering:setup}

\paragraph{Building the steering direction.}
We place each character $c$ at the end of the phonetic, semantic,
and neutral probe prompts and extract its hidden state
in each prompt (Eq.~\ref{eq:probe-hidden}).
We then average these three hidden states:\[
\bar{\mathbf{h}}_c^{(l)}
= \frac{1}{3}\sum_{f\in\{\text{phon},\text{sem},\text{neut}\}}
\mathbf{h}_{c,f}^{(l)}.
\] For a transition
$A\!\to\!B$ (e.g.\ initial \hlinitial{j}$\to$\hlinitial{l}) the
unsupervised steering direction is the mean difference over minimal pairs
that differ \emph{only} in the target dimension:
\begin{equation}
v_{\text{dir}} \;=\; \frac{1}{|\mathcal{P}|}\sum_{(a,b)\in\mathcal{P}}
   \bigl(\bar{\mathbf{h}}_b^{(l)} - \bar{\mathbf{h}}_a^{(l)}\bigr),
\label{eq:steer-dir}
\end{equation}
where $\mathcal{P}$ contains character pairs with source label $A$
and target label $B$, matched on the other two phonological
dimensions. For example, for \hlinitial{j}$\to$\hlinitial{l},
the pair 吉~(\hlinitial{j}\hlfinal{i}\hltone{2}) and
梨~(\hlinitial{l}\hlfinal{i}\hltone{2}) shares the same final
and tone but differs in initial.
The $v_{\text{dir}}$ is usually referred to as the steering vector.

We use an average of 13.7, 9.8, and 144.4 character-pair records per transition
 to construct initial, final, and tone steering vectors, respectively.
By assessing target-class characters not used to construct the vectors
(Appendix~\ref{app:heldout}), we examine the generalization ability of our method.

\paragraph{Injection.}
We add the direction to the residual stream at the source character's position,
$\mathbf{h} \leftarrow \mathbf{h} + \alpha\, v_{\text{dir}}$,
and sweep $\alpha$ upward from $0$ \orange{to $10$}. We intervene at five sites (the embedding and layers $8/13/20/24$) over $82$
transitions ($40$ initial, $30$ final, $12$ tone). The following paragraphs discuss
the embedding and layers $8/13/20$; complete results for all five sites are
provided in Appendix~\ref{app:steer-alpha}.

\paragraph{Measuring the effect.}
For initial, final, and tone transitions, we use parallel prompts:
\emph{“$\langle c\rangle$”的同\{声母/韵母/声调\}字，比如“}
(``a character with the same initial/final/tone as
$\langle c\rangle$, for example''). We sum the next-token probabilities of single-token hanzi in
the target and source classes to obtain $P_\alpha(B)$ and
$P_\alpha(A)$. Their difference,
$a(\alpha)=P_\alpha(B)-P_\alpha(A)$, gives the target--source
advantage. We report:
\begin{equation}
R_{\mathrm{peak}}=\max_\alpha a(\alpha),
\qquad
\Delta_{\mathrm{peak}}
=\max_\alpha[a(\alpha)-a(0)].
\label{eq:steer-effect}
\end{equation}
$R_{\mathrm{peak}}$ is the largest target--source advantage
over the sweep, whereas $\Delta_{\mathrm{peak}}$ measures
its largest increase from $\alpha=0$. Thus, a larger target-source advantage means the feature is easier to steer and vice versa. We report peaks because responses can decline at large
$\alpha$ (Appendix~\ref{app:steer-alpha}).

\section{Steering Is Invariant to the Choice of Template}
\label{app:steer-frame}
Because the steering direction is a difference of bucket-matched minimal pairs
(Eq.~\ref{eq:steer-dir}), the shared extraction template cancels.
Table~\ref{tab:steer-frame} verifies this: rebuilding each direction from the \emph{single}
sentence-final phonetic prompt (``\,{这个字的读音：$\langle c\rangle$}\,'') instead of the
three-template average of \S\ref{sec:steering-setup} changes the per-site peak
target$-$source advantage by at most $0.02$, for both models and all three dimensions.

\begin{table*}[t]
\centering
\small
\caption{Per-site peak target--source advantage under the three-template averaged extraction, with the single phonetic-template extraction in parentheses. Both columns come from a separate re-extraction run, so values can differ slightly from Table~\ref{tab:steer-layer}. The two extractions agree to $\le 0.02$ at every site; the last column is the maximum absolute difference across the five sites. All three dimensions reverse at \layer{24}.}
\label{tab:steer-frame}
\begin{tabular}{llccccc c}
\toprule
Model & Dim & Emb & \layer{8} & \layer{13} & \layer{20} & \layer{24} & max $|\Delta|$ \\
\midrule
7B-base & \initial{} & $0.57\,(0.58)$ & $0.53\,(0.53)$ & $0.51\,(0.51)$ & $0.46\,(0.46)$ & $-0.40\,(-0.39)$ & $0.02$ \\
        & \final{}   & $0.46\,(0.46)$ & $0.40\,(0.39)$ & $0.39\,(0.38)$ & $0.37\,(0.36)$ & $-0.38\,(-0.38)$ & $0.01$ \\
        & \tone{}    & $0.16\,(0.16)$ & $0.12\,(0.12)$ & $0.18\,(0.19)$ & $0.16\,(0.17)$ & $-0.27\,(-0.27)$ & $0.01$ \\
\cmidrule{1-8}
Omni-7B & \initial{} & $0.53\,(0.53)$ & $0.41\,(0.41)$ & $0.37\,(0.37)$ & $0.31\,(0.31)$ & $-0.36\,(-0.36)$ & $0.01$ \\
        & \final{}   & $0.46\,(0.46)$ & $0.36\,(0.34)$ & $0.34\,(0.33)$ & $0.32\,(0.31)$ & $-0.38\,(-0.38)$ & $0.02$ \\
        & \tone{}    & $0.19\,(0.19)$ & $0.14\,(0.14)$ & $0.18\,(0.18)$ & $0.16\,(0.17)$ & $-0.19\,(-0.19)$ & $0.00$ \\
\bottomrule
\end{tabular}
\end{table*}

\section{Steering Dynamics under Increasing Injection Strength}
\label{app:steer-alpha}
\begin{figure*}[t]

  \centering
  \includegraphics[width=0.82\linewidth]{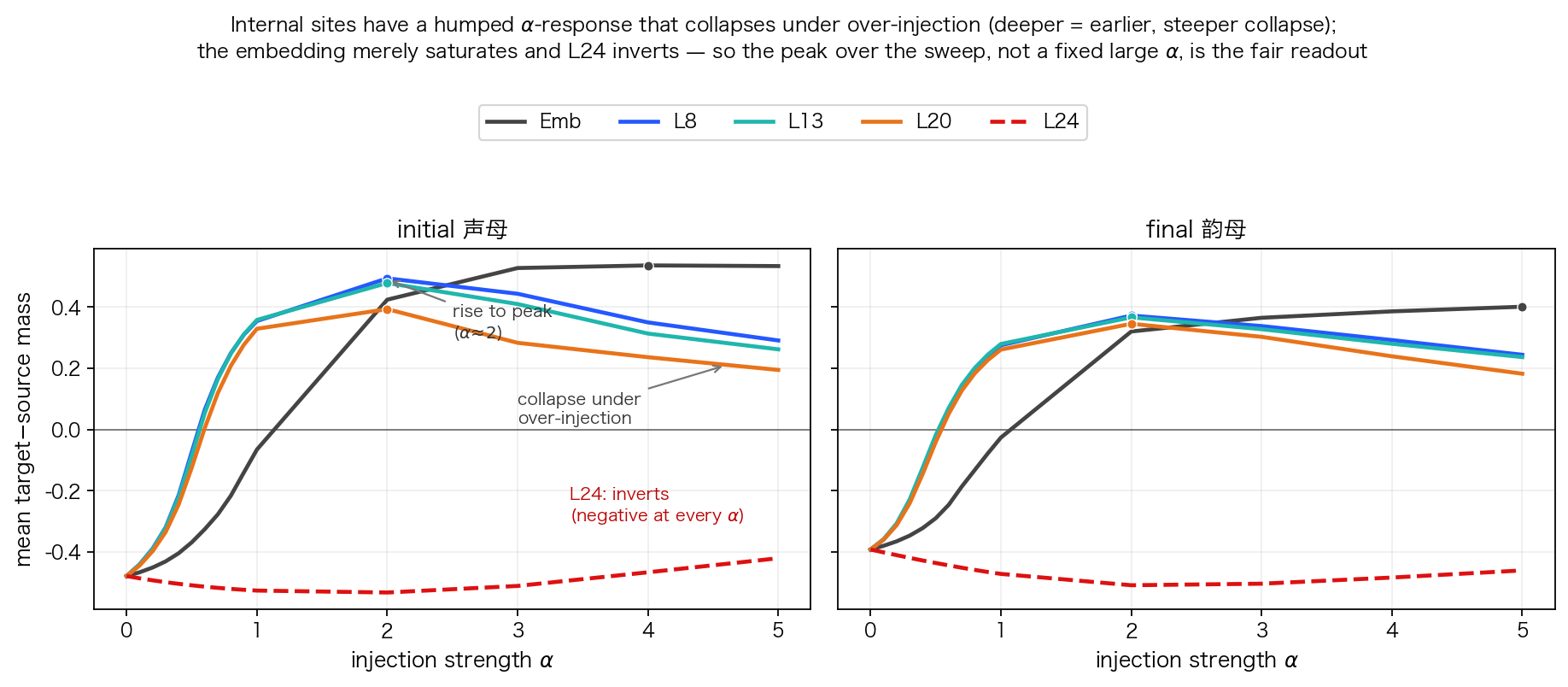}
  \caption{\textbf{Steering $\alpha$-response}, Qwen2.5-7B-base (initial and final transitions
  pooled). Mean target$-$source class mass as a function of the injection strength $\alpha$,
  one curve per site. The embedding saturates; the internal sites peak near $\alpha{\approx}2$
  (dots) and then collapse under over-injection; \layer{24} inverts. The main text reports the
  \emph{peak} of each curve, not its end-point.}
  \label{fig:alpha}
\end{figure*}

Figure~\ref{fig:alpha} shows the per-site response to increasing injection strength
$\alpha$---the basis for the peak-over-sweep scoring of \S\ref{sec:steering-emb}; the
\emph{Collapse} column of Table~\ref{tab:steer-layer} reports the rate of response
decline after the peak.

\section{Strongest Single Transitions}
\label{app:steer-trans}
Table~\ref{tab:steer-trans} lists the strongest single segmental transitions at \layer{8},
the exemplar site of \S\ref{sec:steering-transfer}: both initials and finals produce large,
early flips, while tone transitions are weaker (their ordering is in
\S\ref{sec:steering-graded}).

\begin{table}[h]
\centering
\small
\caption{Strongest single segmental transitions at \layer{8} (peak target$-$source
advantage).}
\label{tab:steer-trans}
\begin{tabular}{llcc}
\toprule
Dim. & Transition & Probe & Peak adv. \\
\midrule
\initial{} & \ipa{tɕ}$\to$\ipa{l} & 吉 & $0.82$ \\
\initial{} & \ipa{p}$\to\varnothing$ & 遍 & $0.77$ \\
\initial{} & \ipa{m}$\to\varnothing$ & 觅 & $0.77$ \\
\initial{} & \ipa{pʰ}$\to$\ipa{l} & 皮 & $0.77$ \\
\final{}   & \ipa{u}$\to$\ipa{ɑʊ} & 簿 & $0.63$ \\
\final{}   & \ipa{jɛn}$\to$\ipa{joʊ} & 莲 & $0.57$ \\
\bottomrule
\end{tabular}
\end{table}

\section{Linear Probing Setup}
\label{app:probing-setup}

\paragraph{Character selection.}
Starting from 2,357 monophonic characters in the
3,500-character pool, we retain phonological classes
with at least 20 examples for stable cross-validation.
This removes five rare finals and 57 characters,
leaving 2,300 characters with 22 initial classes,
31 final classes, and four tone classes.

\paragraph{Representation extraction.}
For a character $c$ presented in prompt $s_c$, we
extract its hidden state at layer $l$:
\begin{equation}
\mathbf{h}_c^{(l)}
=
\operatorname{LM}_{\leq l}(s_c)
\bigl[\operatorname{pos}(c)\bigr]
\in \mathbb{R}^{d},
\label{eq:probe-hidden}
\end{equation}
where $\operatorname{pos}(c)$ denotes the character's
token position and $l=0$ denotes the input embedding.
At layer 0, the representation is the character
embedding itself, so the four prompt conditions
coincide.

\paragraph{Probe training and evaluation.}
For each layer, prompt condition, and phonological
dimension, we standardize the representations and
reduce them to 32 dimensions using PCA.
Both standardization and PCA are fitted only on
the training split of each fold.
We then train an $\ell_2$-regularized multinomial
logistic-regression classifier and evaluate it using
five-fold stratified cross-validation.

We report macro-$F_1$ and selectivity.
For the latter, we repeat probing with a fixed
permutation of the phonological labels and compute
\begin{equation}
\mathrm{Selectivity}
=
F_1^{\mathrm{true}}
-
F_1^{\mathrm{shuffled}}.
\end{equation}

\end{document}